\documentclass[a4paper,fleqn]{cas-sc}

\usepackage[authoryear]{natbib}
\usepackage{amssymb}
\usepackage{amsmath}
\usepackage{CJKutf8}
\usepackage{url}
\usepackage{algorithm}
\usepackage{algorithmic}
 \usepackage{color} 
\usepackage{ulem}
\usepackage{mathrsfs}
\usepackage{graphicx}
\usepackage{makecell}

\newif\ifdraft
\drafttrue

\begin{document}
\let\WriteBookmarks\relax
\def\floatpagepagefraction{1}
\def\textpagefraction{.001}

\newtheorem{example}{\textbf{Example}}

% \shorttitle{Span-level emotion-cause-category triplet extraction}

% \shortauthors{D. Yang, X. Li, Z. Zhao, F. Huang, K. Song}

\title [mode = title]{Span-level Emotion-Cause-Category Triplet Extraction with Instruction Tuning LLMs and Data Augmentation}                      
\author[1]{Xiangju Li}
\ead{xiangjul@sdust.edu.cn}
\author[1]{Dong Yang}
\ead{202283060034@sdust.edu.cn}
\author[1]{Xiaogang Zhu}
\ead{202483060198@sdust.edu.cn}
\author[2]{Faliang Huang}
\ead{hfl@nnnu.edu.cn}
\author[1]{Peng Zhang}
\ead{pengzhang_skd@sdust.edu.cn}
\author[1]{Zhongying Zhao}
\cormark[1]
\ead{zyzhao@sdust.edu.cn}

\affiliation[1]{organization={College of Computer Science and Engineering},
    city={Shandong University of Science and Technology},
    postcode={266500 Qingdao}, 
    country={China}}
    
\affiliation[2]{organization={Guangxi Key Lab of Human-machine Interaction and Intelligent Decision},
    city={Nanning Normal University},
    postcode={530100 Nanning}, 
    country={China}}

% \affiliation[3]{organization={School of Computer Science and Engineering},
%     city={Northeastern University},
%     postcode={110167 Shenyang}, 
%     country={China}}
    
% \affiliation[4]{organization={DAMO Academy},
%     city={Alibaba Group},
%     postcode={310000 Hangzhou}, 
%     country={China}}

% \ead{Email address: internyang_2020@163.com (D. Yang); lixiangju100@163.com (X. Li); zzysuin@163.com (Z. Zhao); faliang.huang@gmail.com (F. Huang); kaisong.sks@alibaba-inc.com}

% % Corresponding author text
% \cortext[cor1]{Corresponding author}
% % \cortext[cor2]{Corresponding author}

% \fntext[fn1]{This is the first author footnote. but is common to third
% \fntext[fn2]{Another author footnote, this is a very long footnote and
%   it should be a really long footnote. But this footnote is not yet
%   sufficiently long enough to make two lines of footnote text.}

% \nonumnote{This note has no numbers. In this work we demonstrate $a_b$
%   the formation Y\_1 of a new type of polariton on the interface
%   between a cuprous oxide slab and a polystyrene micro-sphere placed

\begin{abstract}
Span-level emotion-cause-category triplet extraction represents a novel and complex challenge within emotion cause analysis. This task involves identifying emotion spans, cause spans, and their associated emotion categories within the text to form structured triplets. While prior research has predominantly concentrated on clause-level emotion-cause pair extraction and span-level emotion-cause detection, these methods often confront challenges originating from redundant information retrieval and difficulty in accurately determining emotion categories, particularly when emotions are expressed implicitly or ambiguously. To overcome these challenges, this study explores a fine-grained approach to span-level emotion-cause-category triplet extraction and introduces an innovative framework that leverages instruction tuning and data augmentation techniques based on large language models. The proposed method employs task-specific triplet extraction instructions and utilizes low-rank adaptation to fine-tune large language models, eliminating the necessity for intricate task-specific architectures. Furthermore, a prompt-based data augmentation strategy is developed to address data scarcity by guiding large language models in generating high-quality synthetic training data. Extensive experimental evaluations demonstrate that the proposed approach significantly outperforms existing baseline methods, achieving at least a 12.8\% improvement in span-level emotion-cause-category triplet extraction metrics. The results demonstrate the method’s effectiveness and robustness, offering a promising avenue for advancing research in emotion cause analysis. The source code is available at \url{https://github.com/zxgnlp/InstruDa-LLM}.
% \url{https://github.com/ZZY-GraphMiningLab/InstruDA}.
\end{abstract}

% \includegraphics{figs/grabs.pdf}

% \begin{highlights}

% % \item The span-level emotion-cause-category triplet extraction task brings more sufficient information to the user.
% % \item We Formalize the task as a table-filling problem, which provides a novel perspective to represent triplets.

% % \item We modify a dataset to better fit the span-level emotion-cause-category triplet extraction task.
% % \item The results show the effectiveness of our approach in comparison with baselines.

% % \item We aim to extract span-level emotion-cause-category triplet to offer more precise information.
% \item We extract span-level emotion-cause-category triplet to offer more precise information.
% \item We formalize the task as a table-filling problem and design a table-filling-based approach.
% \item The proposed approach works in an end-to-end manner.
% \item A label-aware mechanism is designed to capture word-candidate label relationships.
% \item A dataset is constructed and the results show the efficacy of our approach.
% \end{highlights}

% Keywords
\let\printorcid\relax
\maketitle

\section{Introduction}

% The rapid advancement of information technology has spurred significant progress in Natural Language Processing (NLP).  Within the flourishing field of NLP, a multitude of specialized tasks \citep{DBLP:journals/eswa/JiaMYNM25, DBLP:journals/eswa/DongW25} have emerged, among which Emotion Cause Analysis (ECA) stands out as a particularly critical area of focus \citep{gu-etal-2024-emoprompt, LI2024121386}. ECA is dedicated to the nuanced analysis of user emotions within textual data, aiming to uncover not only what emotions are expressed but also why they are elicited.  As a fine-grained emotion analysis task \citep{10.5555/1860631.1860637}, ECA has garnered considerable attention from both industry and academia \citep{xia-ding-2019-emotion, CHEN2022107965, ZHU2024111342}. Effectively performing ECA necessitates that models possess a deep understanding of textual semantics and the intricate causal relationships that contribute to emotional expression.  This capability is increasingly recognized as essential for achieving a more profound comprehension of human affective communication \citep{DBLP:journals/kbs/ZhuWTZCW24,DBLP:conf/emnlp/WangTZ24}.

The rapid evolution of information technology has driven remarkable advancements in Natural Language Processing (NLP). Within this expanding domain, numerous specialized tasks \citep{DBLP:journals/eswa/JiaMYNM25, DBLP:journals/eswa/DongW25} have emerged, with emotion cause analysis gaining particular prominence \citep{gu-etal-2024-emoprompt, LI2024121386}.
Emotion cause analysis focuses on the in-depth examination of user emotions in textual data, seeking to determine not only the emotions conveyed but also the underlying reasons behind them. As a fine-grained emotion analysis task \citep{10.5555/1860631.1860637}, emotion cause analysis has attracted significant interest from both academic and industrial communities \citep{xia-ding-2019-emotion, CHEN2022107965, ZHU2024111342}. Successfully conducting emotion cause analysis requires models to develop a deep understanding of textual semantics and the complex causal relationships that shape emotional expressions. This ability is increasingly recognized as a crucial factor in achieving deeper insights into human affective communication \citep{DBLP:journals/kbs/ZhuWTZCW24, DBLP:conf/emnlp/WangTZ24}.

\begin{example}
\label{ext:exam1}
Examples of ambiguous emotion expressions.
\begin{enumerate}
    \item As she accepted the award, her voice \textbf{trembled} with emotion and gratitude.
    \item Facing the angry crowd, his voice \textbf{trembled} as he tried to explain himself.
\end{enumerate}
\end{example}

% Current ECA research primarily focuses on clause-level emotion-cause pair extraction (ECPE) and emotion cause extraction (ECE) tasks. However, these tasks present inherent limitations: (1) Extended clauses often encompass extraneous information irrelevant to emotions or causes, thereby diminishing the precision of clause-level extraction. (2) Ambiguous emotional expressions within the text can lead to inaccuracies in emotion category extraction. For instance, as illustrated in Example~\ref{ext:exam1}, the expression \textbf{\textit{trembled}} might signal happiness for an individual accepting an award with gratitude, yet denote fear for someone facing an angry crowd while trying to explain themselves. Therefore, our research centers on the fine-grained \textbf{s}pan-level \textbf{e}motion-\textbf{c}ause-\textbf{c}ategory triplet \textbf{e}xtraction (SECCE) task to address these shortcomings and achieve more precise and nuanced emotion analysis. By employing fine-grained spans as extraction units, this task facilitates more accurate localization of pertinent emotion spans within clauses and minimizes interference from superfluous information. Furthermore, the incorporation of emotion categories enables a more precise characterization of the expressed emotion.  The SECCE task demands that models possess a profound understanding of textual semantics, the causal mechanisms behind emotions, and the ability to discern emotion categories, rendering it exceptionally challenging.

Current emotion cause analysis research predominantly focuses on clause-level tasks such as emotion-cause pair extraction and emotion cause extraction. However, these approaches come with inherent drawbacks:
(1) Extended clauses often contain extraneous details unrelated to emotions or their causes, reducing the accuracy of clause-level extraction.
(2) Ambiguous emotional expressions within the text can lead to misinterpretations of emotion categories. For instance, as shown in Example \ref{ext:exam1}, the word \textit{trembled} might indicate joy for someone receiving an award with gratitude, yet signify fear for an individual attempting to explain themselves to an angry crowd.
To overcome these limitations, this study focuses on the span-level emotion-cause-category triplet extraction, a fine-grained approach designed to enhance precision in emotion analysis. By using spans rather than full clauses as extraction units, span-level emotion-cause-category triplet extraction allows for more accurate identification of relevant emotion spans while minimizing interference from irrelevant content. Additionally, incorporating emotion categories provides a clearer and more precise characterization of the emotions being expressed.

\begin{figure}[ht]
\centering
\includegraphics[width=1.0\columnwidth]{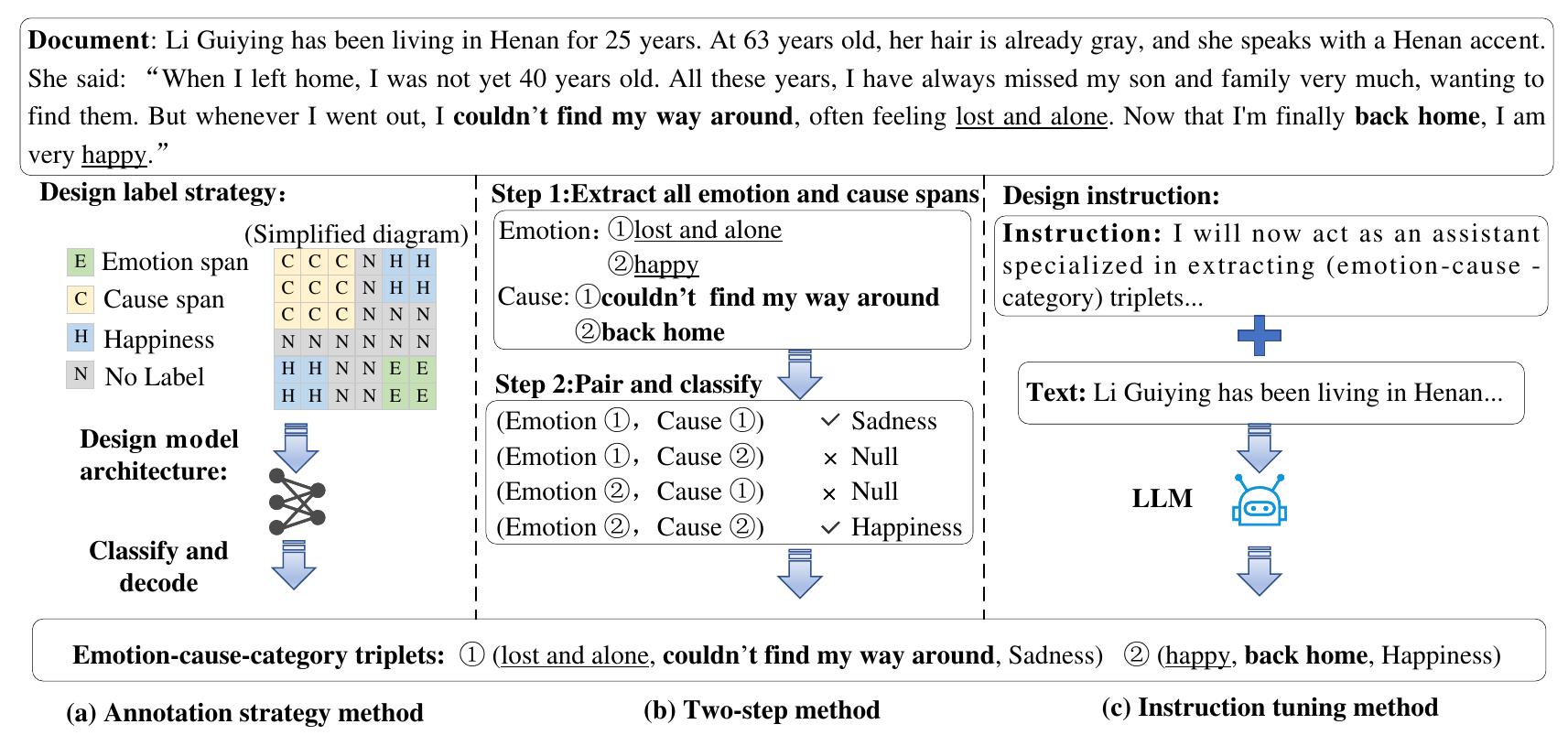} 
\caption{Three methods for extracting emotion-cause-category triplets.}
\label{example}
\end{figure}

The span-level emotion-cause-category triplet extraction task demands a deep understanding of textual semantics, the causal mechanisms underlying emotions, and the ability to accurately classify emotion categories, making it a particularly complex and challenging problem. 
Although span-level emotion-cause-category triplet extraction holds significant potential for advancing emotion analysis, research in this domain remains limited. The only existing method by \cite{YANG2025126062} adopts a table-filling paradigm Figure \ref{example} (a) that encodes emotion-cause relationships through a two-dimensional table format. However, this approach rely heavily on predefined annotation schemes and custom specialized model architectures, thereby increasing implementation complexity and restricting the method’s adaptability across diverse applications.
An alternative two-step methodology Figure \ref{example} (b) first employs sequence labeling to detect emotion and cause spans, then pairs these spans and classifies emotion categories. However, this method is prone to error propagation, a common drawback in multi-stage pipelines. Errors in the initial extraction of emotion and cause spans can negatively impact the classifier’s performance in the next step, ultimately reducing the accuracy of the final triplets. Given the inherent weaknesses of both table-filling and two-step approaches, particularly their susceptibility to error propagation and limited generalizability, this study aims to explore more robust and efficient methodologies to enhance span-level emotion-cause-category triplet extraction performance.

In contrast to the preciously discussed methods, this study introduces an end-to-end instruction tuning approach using large language models (LLMs), as depicted in Figure \ref{example} (c). By designing structured instruction templates, the proposed method employs the advanced language comprehension and logical reasoning capabilities of LLMs to generate pre-formatted triplet text. This end-to-end framework inherently mitigates error propagation, addressing a major drawback of multi-stage approaches. Additionally, instruction-based fine-tuning of LLMs eliminates the need for complex, task-specific model architectures, thereby enhancing adaptability across different tasks. A key advantage of this approach is its ability to directly produce triplet text, bypassing the need for additional decoding steps and streamlining the data processing pipeline. As shown in Figure~\ref{example} (c), the model’s output can be efficiently processed programmatically, simplifying the statistical analysis of experimental metrics. To further optimize computational efficiency, this study incorporates low-rank adaptation (LoRA) \citep{DBLP:conf/iclr/HuSWALWWC22} techniques, significantly reducing the resource demands typically associated with fine-tuning large models. However, the effectiveness of LLM-based approaches is often contingent on the availability of extensive training data, necessitating additional strategies to address data scarcity challenges.

To overcome the challenge of data scarcity, which is particularly relevant for intricate tasks such as emotion-cause-category triplet annotation, this study further proposes the use of LLMs for data augmentation. Existing Chinese emotion cause analysis datasets \citep{GuiWXLZ16} are often limited in size, largely due to the complexity and time-intensive nature of triplet annotation. To address this constraint and maximize the effectiveness of the instruction-tuning approach, LLMs are utilized to generate additional training data. By harnessing their advanced text generation capabilities, structured prompt templates are designed to guide the model in producing high-quality samples that adhere to the emotion-cause-category triplet format. This strategy not only expands the dataset but also enhances the diversity of training samples.

To systematically evaluate the effectiveness of the proposed approach, which integrates both instruction tuning and data augmentation, a series of comparative experiments were conducted against multiple baseline methods. The results indicate that the instruction tuning strategy yields substantial performance improvements, while the LLM-based data augmentation technique effectively expands the training dataset, further enhancing model performance. In summary, the key contributions of this study are as follows:

\begin{itemize}
    \item An instruction tuning approach leveraging large language models is proposed, enabling the generation of emotion-cause-category triplet through carefully designed instruction templates.
    \item Large language models are utilized for data augmentation, where prompt templates are employed to guide the generation of standardized training samples, thus enhancing the available data resources for model training.
    \item Span-level emotion-cause-category triplet extraction experiments are carried out on a benchmark emotion cause analysis dataset, accompanied by a set of auxiliary task experiments. The results demonstrate the effectiveness of the proposed approach.
\end{itemize}

\section{Related work} \label{Related work}
\subsection{Instruction tuning}
% Instruction tuning aims to fine-tune large language models on diverse natural language datasets, enabling the models to better understand and execute human instructions. Designing task-specific instructions can improve the adaptability of large language models to specific tasks while simplifying the process of designing complex network structures and training pipelines for each new task. In recent years, instruction tuning methods have shown outstanding performance in various NLP tasks \citep{10.5555/3455716.3455856, DBLP:conf/acl/DingL0024, DBLP:conf/emnlp/ZhangD0O24, DBLP:conf/acl/ZhangZWZ24}. Instruction tuning has seen limited application in the field of emotion cause analysis, whereas in the related field of aspect-based sentiment analysis, some research has been conducted.
Instruction tuning focuses on fine-tuning large language models using diverse natural language datasets, enhancing
their ability to comprehend and respond to human instructions. Designing task-specific instructions improves the
adaptability of these models to particular tasks, which also simplifies the need for developing complex network
architectures and training pipelines for each new task. Recently, instruction tuning has demonstrated impressive results across various NLP applications \citep{10.5555/3455716.3455856, DBLP:conf/acl/DingL0024, DBLP:conf/emnlp/ZhangD0O24, DBLP:conf/acl/ZhangZWZ24}. While its use in emotion cause analysis remains limited, it has attracted scholars in the related domain of aspect-based sentiment analysis.

% \cite{varia-etal-2023-instruction} proposed an instruction tuning framework that transforms multiple subtasks in aspect-based sentiment analysis into question-answering formats, using the T5 model for instruction tuning, significantly improving the model's performance in few-shot scenarios. Since existing methods have not fully explored how to effectively leverage the knowledge of large language models to handle implicit aspects and opinions, \cite{zhang-etal-2024-instruction} combined instruction tuning with supervised contrastive learning to improve the model's accuracy in predicting aspect sentiment quadruples, enhancing the model's ability to model implicit aspects and opinions in text. To address the limitations of fixed examples in aspect-based sentiment analysis tasks, \cite{zheng-etal-2024-instruction} proposed a retrieval-based example ranking instruction learning method. This method uses a retriever to select examples to form instruction templates and scores and reasons about the examples using large language models, thereby improving the model's generation efficiency and performance.

\cite{varia-etal-2023-instruction} introduced an instruction tuning framework that reformulates multiple subtasks in aspect-based sentiment analysis into question-answering formats, employing the T5 model for instruction tuning. This approach significantly enhanced the model's performance, particularly in few-shot learning scenarios. Since existing methods have not fully explored how to effectively utilize large language models for handling implicit aspects and opinions, \cite{zhang-etal-2024-instruction} integrated instruction tuning with supervised contrastive learning to improve the model's accuracy in predicting aspect sentiment quadruples, thereby enhancing its ability to capture implicit aspects and opinions in text. To overcome the limitations of fixed examples in aspect-based sentiment analysis tasks, \cite{zheng-etal-2024-instruction} proposed a retrieval-based example ranking instruction learning method. This approach employs a retriever to select relevant examples for forming instruction templates, while large language models score and reason about these examples, thus improving generation efficiency and model performance.

\subsection{Text data augmentation}
% Large language models (LLMs) have recently emerged as a powerful tool in text data augmentation, demonstrating notable performance gains \citep{NEURIPS2020_1457c0d6}.  Researchers have begun to explore the distinctions between LLM-based and traditional data augmentation techniques. For instance, \cite{piedboeuf-langlais-2023-chatgpt} employed ChatGPT to augment data by paraphrasing sentences, generating semantically similar yet lexically diverse variations \citep{piedboeuf-langlais-2023-chatgpt}.  In another approach, \cite{yoo-etal-2021-gpt3mix-leveraging} utilized LLMs to predict soft labels and subsequently applied knowledge distillation, combined with text perturbation methods like lexical substitution and word order shuffling, to create augmented datasets.  Addressing the challenge of adapting general instructions to diverse downstream tasks, \cite{li-etal-2024-empowering} leveraged LLMs to generate a spectrum of augmented instructions. They further implemented a scoring mechanism to select task-pertinent instructions, thereby enhancing model generalization \citep{li-etal-2024-empowering}.  Moreover, \cite{chen-etal-2024-minprompt} focused on few-shot question answering, employing graph algorithms and unsupervised question generation to extract the most informative textual data, thus minimizing the reliance on extensive fine-tuning data and improving training efficiency.
Large language models have recently emerged as a powerful tool for augmenting text data, resulting in substantial improvements in performance \citep{NEURIPS2020_1457c0d6}. In this context, numerous investigations have been conducted focusing on the differences between LLM-based augmentation and traditional techniques. For instance,  \cite{piedboeuf-langlais-2023-chatgpt} utilized ChatGPT to enhance datasets by paraphrasing sentences, generating variations that were semantically similar but lexically diverse. \cite{yoo-etal-2021-gpt3mix-leveraging} applied LLMs to predict soft labels and then used knowledge distillation alongside text perturbation methods such as lexical substitution and word order shuffling to create augmented datasets. Addressing the challenge of adapting general instructions to a variety of downstream tasks, \cite{li-etal-2024-empowering} employed LLMs to generate a wide range of augmented instructions and implemented a scoring system to select task-relevant instructions, which improved model generalization. Additionally, \cite{chen-etal-2024-minprompt} focused on few-shot question answering, using graph algorithms and unsupervised question generation to extract the most informative data, thereby reducing the need for extensive fine-tuning data and enhancing training efficiency.

Similarly, large language models can be employed to augment data for the span-level emotion-cause-category triplet extraction task. This can be achieved by rephrasing the original text to create new variations that maintain semantic consistency while presenting the information differently. However, a key challenge lies in ensuring that the newly generated data aligns with the specific formatting requirements for span-level emotion-cause-category triplets.
% Similarly, the data for the SECCE task can also be augmented using large language models. Specific methods include rewriting the original text to generate new semantically consistent but phrased differently. However, ensuring that the newly generated data adheres to the specific format requirements of span-level emotion-cause-category triplets presents a challenging research problem.

\subsection{Emotion cause analysis}
% ECA aims to identify and extract the causes and emotions in text. As the field has developed, several tasks have been proposed, including emotion cause extraction, emotion-cause pair extraction, and span-level emotion-cause-category triplet extraction.
Emotion cause analysis focuses on identifying and extracting emotions and their corresponding causes from text. Over time, the field has evolved to include various tasks, such as emotion cause extraction, emotion-cause pair extraction, and span-level emotion-cause-category triplet extraction.

The task of emotion cause extraction focuses on identifying the causes of emotions in text. It was first introduced
by \cite{10.5555/1860631.1860637}, who employed a rule-based approach to detect keywords that signify causes behind emotions. Later, \cite{chen-etal-2010-emotion} demonstrated that clause-level granularity is more effective for emotion cause extraction and proposed a multi-label method to identify emotion-cause clauses. In response to the limited availability of publicly accessible datasets, \cite{gui-etal-2016-event} developed an emotion cause extraction dataset based on Sina News, which has become a valuable resource for subsequent research. To fully exploit the structural information in clause dependency relationships, \cite{HU2021106584} constructed a dependency graph between clauses and applied graph convolutional networks to capture the semantic and structural relationships. Moreover, \cite{yan-etal-2021-position} highlighted that many emotion cause extraction models rely on the relative position of clauses, which limits their ability to generalize when position information is unclear. To address this limitation, \cite{yan-etal-2021-position} incorporated a commonsense knowledge base to enhance the semantic dependency between emotion clauses and potential cause clauses, reducing the impact of position bias.
% The emotion cause extraction task extracts the causes of a given emotion from text. This task was first proposed by \cite{10.5555/1860631.1860637}, who proposed a rule-based method to detect cause keywords behind emotions. \cite{chen-etal-2010-emotion} suggested that clause-level granularity is more suitable for extracting emotion causes and proposed a multi-label method to detect emotion cause clauses. To address the lack of publicly available datasets, \cite{gui-etal-2016-event} constructed an emotion cause extraction dataset based on Sina News, providing an important data resource for subsequent research. To fully utilize the structural information in clause dependency relationships, \cite{HU2021106584} constructed a dependency graph between clauses and used graph convolutional networks to learn the semantic and structural relationships between clauses. \cite{yan-etal-2021-position} pointed out that most existing emotion cause extraction models use the relative position information of clauses for emotion cause extraction, leading to poor generalization ability when processing data with unclear position information. To address this issue, \cite{yan-etal-2021-position} used a commonsense knowledge base to enhance the semantic dependency relationships between candidate clauses and emotion clauses, thereby mitigating position bias.

Emotion cause extraction often depends on emotion annotation before extracting the corresponding causes, which limits its practical application in real-world scenarios. To address this, \cite{xia-ding-2019-emotion} introduced the emotion-cause pair extraction task and proposed a two-step method. The approach first conducts emotion extraction and cause extraction separately through multi-task learning and then pairs and classifies the extracted emotion and cause clauses. To overcome the lack of explicit modeling of information interaction and improve interpretability in existing methods, \cite{9457144} framed the emotion-cause pair extraction task as a sequence labeling problem, introducing a labeling scheme that includes emotion-cause distance information. This framework uses multi-task learning to leverage the correlations between tasks and improve model performance. In response to the challenges faced by multi-task learning approaches in modeling specific features, interaction features, and label prediction inconsistencies, \cite{chen-etal-2022-joint} proposed an alignment mechanism that aligns the features for emotion extraction, cause extraction, and emotion-cause pair extraction tasks, ensuring label consistency through a cross-task alignment mechanism. To address the difficulty in distinguishing emotion-cause pairs associated with different emotion types, \cite{gu-etal-2024-emoprompt} introduced a prompt fine-tuning method for large language models. This method incorporates cause clauses and emotion-type knowledge into the prompt templates and expands emotion labels using an external emotion lexicon. Furthermore, to address the class imbalance issue in this task, \cite{LI2024121386} introduced a summary document extraction task, which selects core clauses from the original document as candidate emotion and cause clauses and then jointly extracts emotions, causes, and emotion-cause pairs from these candidates.

In the emotion-cause pair extraction task, the extraction of long clauses often leads to the inclusion of redundant information that is unrelated to emotions or causes. Additionally, solely focusing on extracting emotional expressions from the raw text may not fully capture the subtle nuances of emotion categories. To address these challenges, the focus shifts to the more refined span-level emotion-cause-category triplet extraction task. While this task is crucial for advancing effective computing systems, it remains a relatively under-explored area with limited existing research. In this emerging field, \cite{YANG2025126062} proposed a table-filling strategy for the span-level emotion-cause-category triplet extraction task. They introduced a label-aware approach to merge emotion and cause information, enabling end-to-end extraction of emotion-cause-category triplets. However, this method requires the design of task-specific model architectures for the table structure. As the annotation scheme and corresponding model architecture must be reconstructed for different task scenarios, this results in high implementation costs. To overcome the need for specialized model architectures, instruction-based fine-tuning of large language models is utilized, with the integration of LoRA to enhance computational efficiency.
% In the ECPE task, extracting long clauses often includes redundant information unrelated to emotions or causes. Furthermore, focusing solely on extracting emotional expressions directly from the raw text may not accurately represent the nuanced emotion categories conveyed. Therefore, we focus on the more fine-grained span-level emotion-cause-category triplet extraction task. Despite its significance in advancing affective computing architectures, this research direction remains an emerging domain with scarce existing literature. In this nascent field, \cite{YANG2025126062} engaged in initial exploration and proposed a table-filling strategy for the SECCE task, designing a label-aware strategy to fuse emotion and cause information, thereby achieving end-to-end extraction of emotion-cause-category triplets. However, this method requires designing specific model architectures for the table structure, and in different task scenarios, the annotation scheme and corresponding model structure need to be reconstructed, resulting in relatively high implementation costs. Therefore, to address the need for specialized model architectures, we employ instructions to fine-tune large language models, utilizing LoRA to enhance efficiency.
%% Use \subsubsection, \paragraph, \subparagraph commands to 
%% start 3rd, 4th and 5th level sections.
%% Refer following link for more details.
%% https://en.wikibooks.org/wiki/LaTeX/Document_Structure#Sectioning_commands

\section{Methodology}

\begin{figure}[ht]
\centering
\includegraphics[width=0.9\columnwidth]{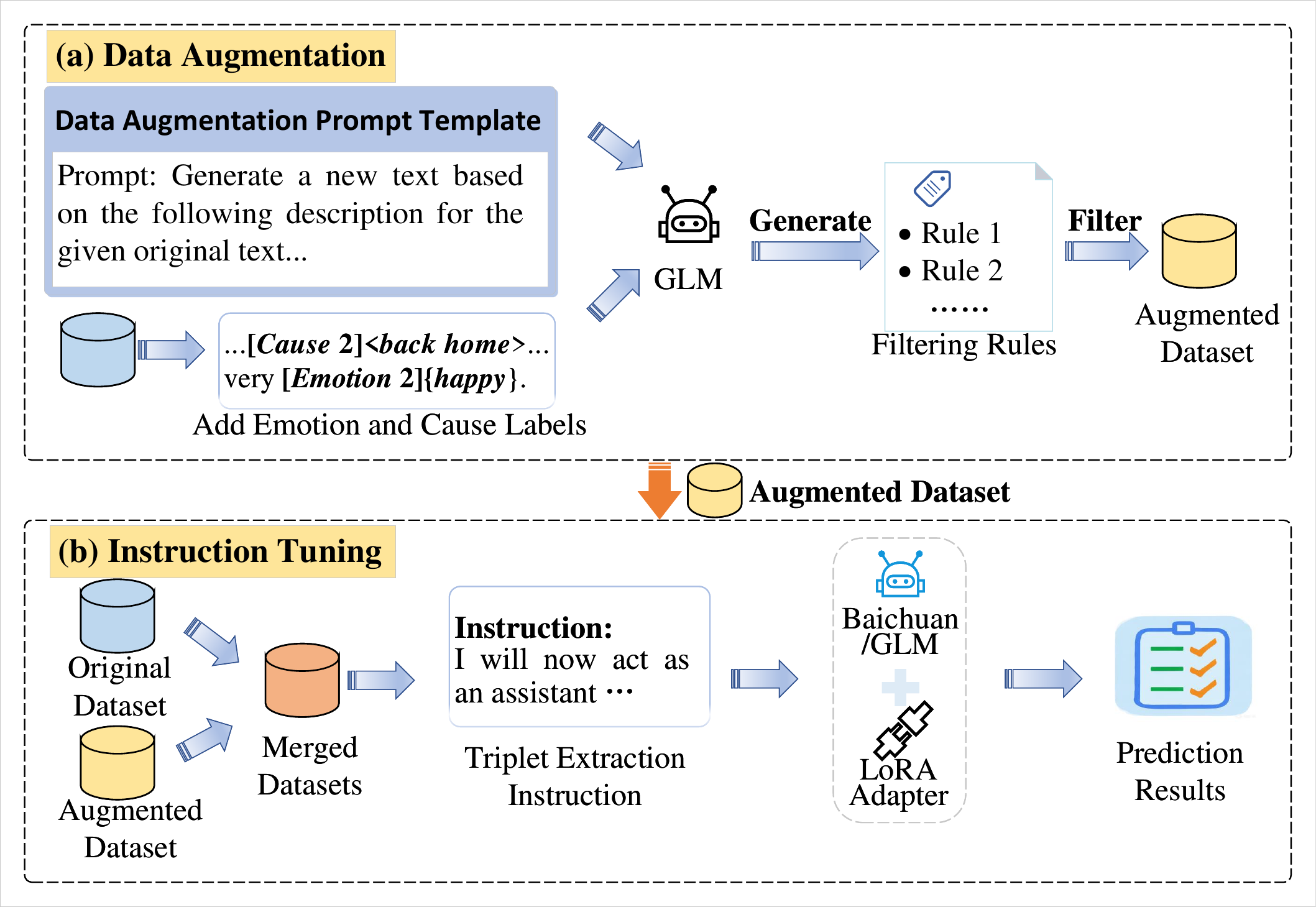} 
\caption{Overview of the proposed model, which mainly consists of two parts: (a) data augmentation and (b) instruction tuning. The data augmentation module generates an augmented dataset, which is then used for instruction tuning to predict emotion-cause-category triplets.}
\label{model}
\end{figure}

% The proposed model for emotion-cause-category triplet extraction based on instruction tuning and data augmentation is shown in Figure~\ref{model}. The model consists of two main modules: the data augmentation module and the instruction tuning module. The data augmentation module aims to generate high-quality augmented training data. Specifically, to help the model accurately locate and identify the elements of the triplet in the original data, an auxiliary labeling strategy is designed to annotate emotion spans and cause spans in the original text. With data augmentation prompts designed, we input the annotated original text and prompts into the model, guiding it to generate new data. In the final step, generated data is filtered for standard-compliant instances and then integrated into the original dataset to enhance the training dataset. In the instruction tuning module, emotion-cause-category triplet extraction instructions are designed to guide the model in extracting span-level emotion-cause-category triplets from the input text. LoRA-based fine-tuning is then performed to reduce training resource consumption and enhance training efficiency.
The model proposed for emotion-cause-category triplet extraction, which combines instruction tuning and data
augmentation, is illustrated in Figure~\ref{model}. This model consists of two primary components: the data augmentation module and the instruction tuning module. The data augmentation module is responsible for generating high-quality augmented training data. Specifically, it includes an auxiliary labeling strategy designed to annotate emotion spans and cause spans in the original text. These annotations help the model accurately locate and identify the elements of the triplet within the input data. After applying the data augmentation prompts, the annotated original text is fed into the model, which generates new data. In the final stage, the generated data is filtered for instances that comply with the required standards and then incorporated into the original dataset to enhance the overall training set. The instruction tuning module focuses on guiding the model to extract span-level emotion-cause-category triplets from the input text. To accomplish this, specific instructions for emotion-cause-category triplet extraction are designed. The model is then fine-tuned using LoRA \citep{DBLP:conf/iclr/HuSWALWWC22}, a technique that helps reduce the consumption of training resources while improving training efficiency.

\subsection{Data augmentation}
\begin{figure}[ht]
\centering
\includegraphics[width=0.95\columnwidth]{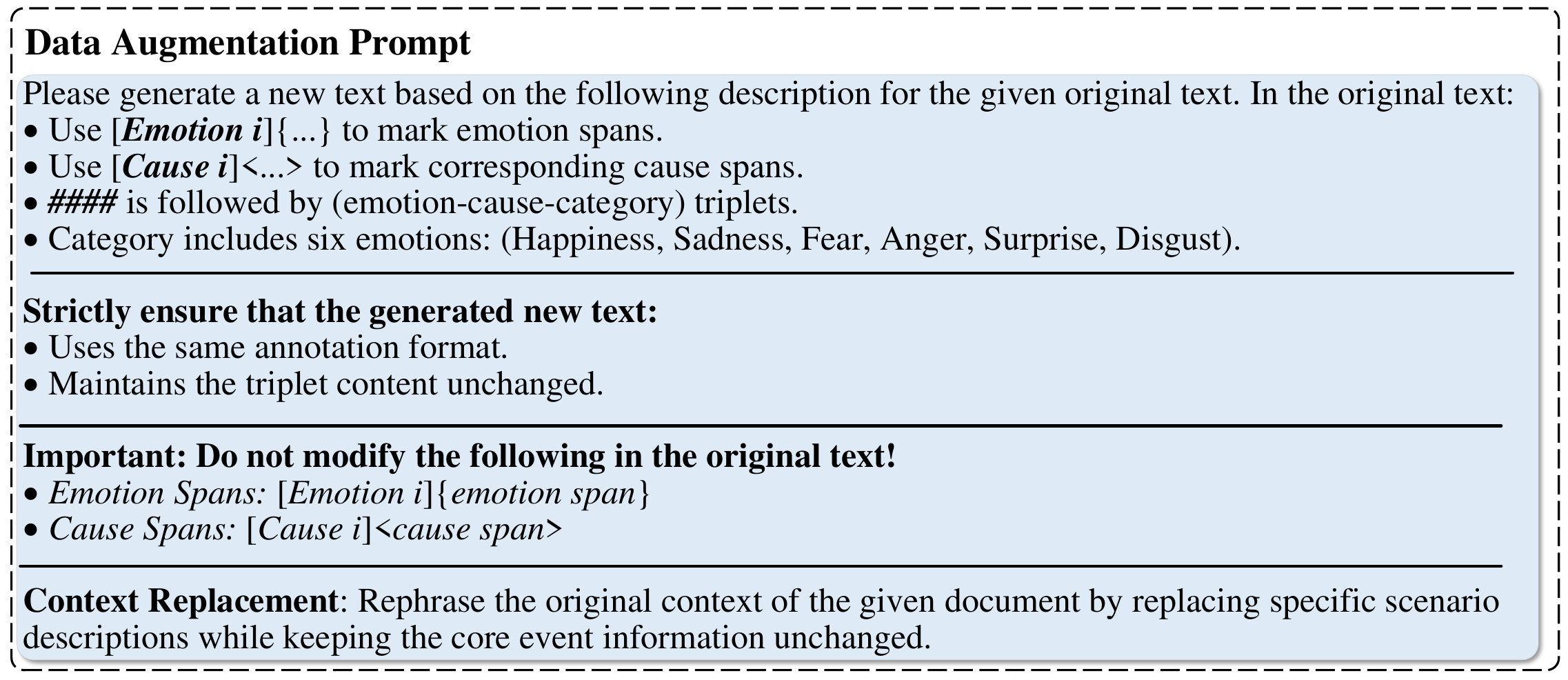} 
\caption{Data augmentation prompt.}
\label{data aug}
\end{figure}

% Data augmentation, a technique for generating synthetic training instances from existing data, is employed in this study.  Harnessing the advanced contextual understanding and text generation capabilities of large language models, we can effectively augment datasets and address the challenge of data scarcity.  For the purpose of ensuring triplet standardization in the generated data, it is stipulated in this study that triplets in the newly generated text must exhibit consistency with those in the original text.  To assist the model in identifying emotion and cause spans in the original text, and consequently, to prevent modifications to these critical components during the generation process, an auxiliary labeling structure is devised.
In this study, data augmentation is utilized as a technique to create synthetic training examples from existing data.
By leveraging the advanced contextual comprehension and text generation abilities of large language models, datasets
can be effectively augmented, addressing the challenge of limited data availability. To maintain consistency with the
original data, the generated triplets must adhere to the same structure and format as those in the source text. To aid
the model in accurately identifying the emotion and cause spans in the original text and to prevent alterations to
these essential components during the generation process an auxiliary labeling framework is developed. As shown in
Figure~\ref{data aug}, the data augmentation prompt establishes a predefined labeling structure. For each input text that contains emotion-cause-category triplets, the corresponding triplets are appended to the end of the text. The labeling process involves marking the emotion and cause spans within each triplet directly in the text. Specifically, for the \textit{i}-th triplet, the emotion span is labeled using the format ``[\textit{Emotion} \textit{i}]\{\textit{emotion span}\}'' and the cause span is labeled with ``[\textit{Cause} \textit{i}]\textless \textit{cause span}\textgreater''. As demonstrated in Figure~\ref{data aug} with the example triplet ``[happy, back home, Happiness]'', the emotion span ``happy'' is marked as ``[\textit{Emotion} 2]\{\textit{happy}\}'' and the cause span ``back home'' as ``[\textit{Cause} 2]\textless \textit{back home}\textgreater''.

% As depicted in Figure~\ref{data aug}, our data augmentation prompt predefines the labeling structure.  For each input text associated with emotion-cause-category triplets, we concatenate these triplets to the end of the text.  The labeling process then involves marking the emotion and cause spans within each triplet directly within the text.  Specifically, for the \textit{i}-th triplet, the emotion span is annotated using the format ``[\textit{Emotion} \textit{i}]\{emotion span\}'' and the cause span with ``[\textit{Cause} \textit{i}]\textless cause span\textgreater ''.  As illustrated in Figure~\ref{data aug} with the example triplet ``[\textit{happy}, \textit{back home}, \textit{Happiness}]'', the emotion span ``\textit{happy}'' is annotated as ``[\textit{Emotion} 2]\{\textit{happy}\}'' and the cause span ``\textit{back home}'' as ``[\textit{Cause} 2]\textless \textit{back home}\textgreater''.
% Auxiliary labeling enables the identification of emotion and cause spans in the original text. Subsequently, the prompt constrains the model to preserve the content of these labeled spans, preventing the generation of non-standard triplets.  Furthermore, this study incorporates a context replacement strategy within the data augmentation prompt to guide text generation, specifically by reconstructing the context while maintaining the event's core information.
Auxiliary labeling facilitates the identification of emotion and cause spans in the original text. Following this, the prompt ensures the model preserves the content of these labeled spans, thereby preventing the generation of non- standard triplets. Additionally, this study introduces a context replacement strategy within the data augmentation prompt, guiding the model to reconstruct the surrounding context while preserving the core information related to the event.

In the final phase of data augmentation, a rule-based filtering process is applied to the generated data to address
potential issues with non-standard triplets that may arise from large language models, ensuring the quality of the
augmented data. This filtering step ensures that selected samples comply with task requirements. Specifically, the text spans must be contained within a single clause and cannot span across multiple clauses. Furthermore, the emotion and cause spans within the augmented triplets must adhere to the pre-established format criteria. The emotion categories within the triplets are also required to align with the six predefined categories. Once filtered, the augmented data is integrated with the original dataset to enhance the training set.
% In the final stage of data augmentation, rule-based filtering is applied to the generated data to mitigate potential non-standard triplet content arising from large language models and ensure data quality. This filtering process selects samples meeting task requirements. Specifically, the span of text spans must be within a single clause segment and not span multiple clauses.  Additionally, emotion and cause spans within augmented triplets must adhere to pre-defined format criteria. Furthermore, emotion categories within triplets must belong to the specified six categories. Subsequently, the filtered augmented data is integrated with the original dataset to augment the training dataset.

\subsection{Instruction tuning}
% Instruction tuning effectively guides large language models to generate emotion-cause-category triplets in a defined format, thus enhancing model performance on the triplet extraction task. To minimize training parameters, we employ low-rank adaptation, fine-tuning only a subset of key parameters to reduce computational resource consumption.
Instruction tuning effectively guides large language models to generate emotion-cause-category triplets in a specified format, thereby improving the model's performance on the triplet extraction task. To minimize the number of trainable parameters, low-rank adaptation is utilized, focusing on fine-tuning only a select subset of crucial parameters, which helps reduce computational resource usage.

\subsubsection{Instruction design}
\begin{figure}[ht]
\centering
\includegraphics[width=0.98\columnwidth]{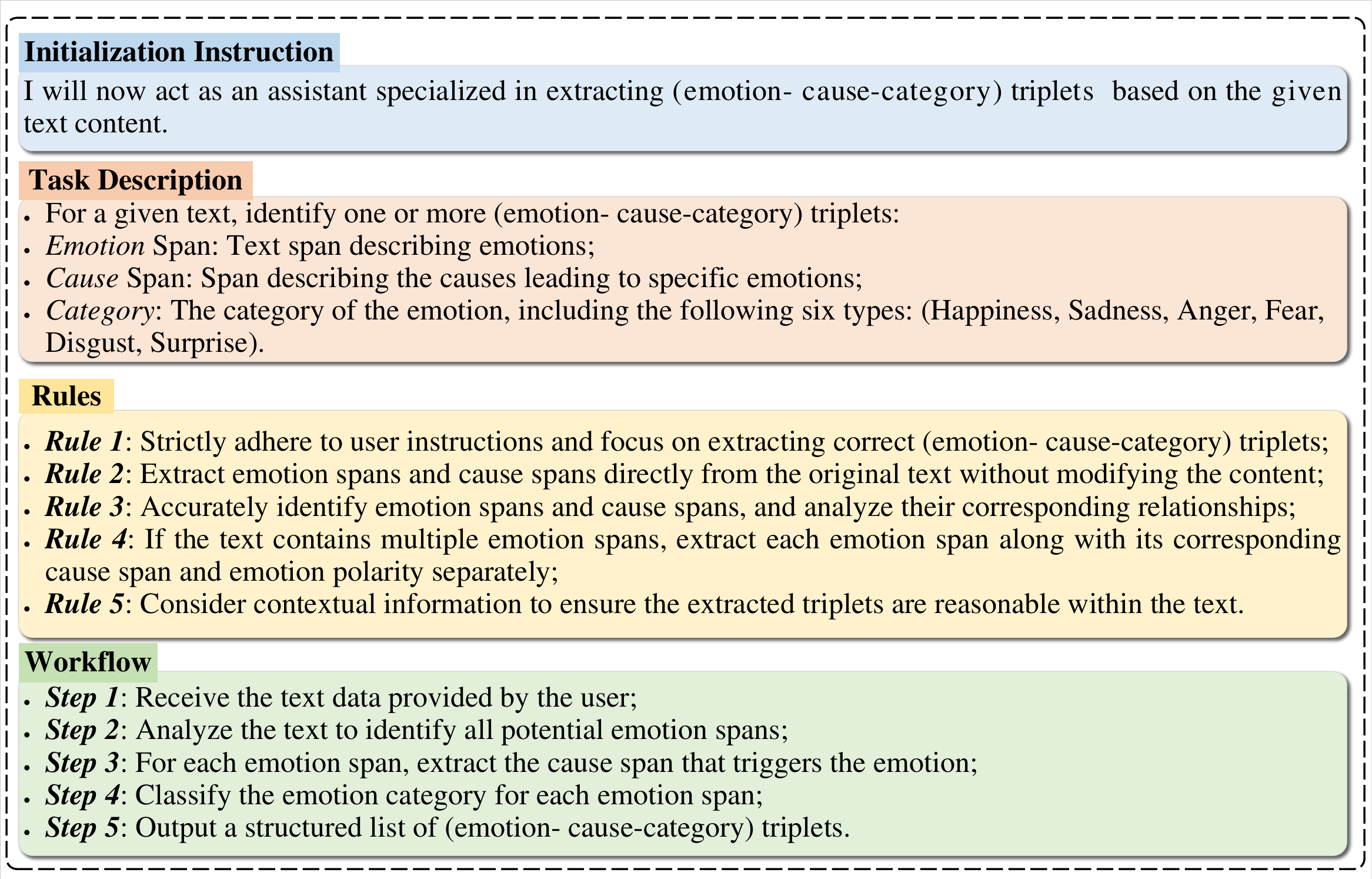} 
\caption{Triplet extraction instruction.}
\label{instruction}
\end{figure}

The model initialization instructions section of the proposed dialogue-style instruction template, as shown in Figure~\ref{instruction}, is designed to specify the task details and constraints, assisting the model in understanding the task at the outset. This template follows a three-part structure: it begins with model initialization instructions (as illustrated in Figure~\ref{instruction}), followed by the input text data, and concludes with the output of emotion-cause-category triplets. This structured approach is intended to guide the model in generating the desired output results. The model initialization instruction defines the model's role and the specific task it is to perform. To provide clarity on the task's purpose and the structure of the triplets, a task description module is incorporated. This module outlines the span-level emotion- cause-category triplet extraction process and provides explicit definitions for each triplet element. Additionally, a rule module is implemented to ensure strict adherence to the task definition during extraction. This module includes five specific rules that guide the model in accurately identifying and extracting the required triplet components.
% The model initialization instructions section of our dialogue-style instruction template is shown in detail in Figure~\ref{instruction}. This section is designed to define specific task specifications and limitations, thus helping the model to initially understand the task.  Our dialogue-style instruction template follows a three-part structure: starting with model initialization instructions (detailed in Figure~\ref{instruction}), followed by input text data, and concluding with output emotion-cause-category triplets.  With this three-part dialogue approach, we expect the model to generate desired output results.

\textit{Rule 1} establishes the foundation by ensuring the model strictly follows the user instructions, confining its actions to the specified operations within the defined workflow. This rule also emphasizes that the output should focus solely on relevant triplet extraction, excluding any unnecessary information. Given that large language models may experience interpretive drift, where the understanding subtly shifts, this can lead to non-compliant spans, such as cause spans extending beyond clause boundaries. To prevent this issue, \textit{Rule 2} explicitly prohibits altering the original text during extraction. Based on this controlled extraction approach, \textit{Rule 3} harnesses the model's inherent capabilities to recognize emotion-cause relationships, improving the accuracy of emotion-cause span pair identification and their relational mapping. Expanding upon this, \textit{Rule 4} guides the model to recognize and link multiple emotion and cause spans along with their associated emotion categories, refining triplet accuracy and enhancing the handling of complex mappings. Finally, \textit{Rule 5} ensures that extracted triplets maintain contextual coherence, guaranteeing that the triplets are reasonable and aligned with the overall context of the text.

% To operationalize these principles, a workflow module is implemented to guide text processing and triplet output in a pre-defined format. Within this module, the model systematically executes a sequence of operations. First, upon receiving user-provided text data, the module initiates emotion span identification. Second, for each identified emotion span, the system proceeds to extract the corresponding cause span that triggers the emotion. Third, the emotion category for each emotion span is classified. Finally, the module outputs a list of emotion-cause-category triplets.  To ensure efficient data utilization, the workflow further specifies a formatted triplet output, which streamlines subsequent data processing, provides clear operational guidance, and minimizes ambiguity in each step.
% Consistency between input and output is critical for the operation. Consequently, the second dialogue is designed to provide pre-defined formatted input to the model, adhering to a fixed format. Upon processing this input, the model directly generates analysis results, presenting them in the third dialogue as a list of emotion-cause-category triplets, thus ensuring a predictable and readily interpretable output structure.
To implement these principles, a workflow module is designed to systematically guide the text processing and
triplet output in a pre-defined format. The module follows a sequence of operations to ensure the extraction process is methodical and efficient. Initially, the model receives user-provided text data, triggering the identification of emotion spans. Once emotion spans are identified, the system proceeds to extract the corresponding cause spans that trigger the emotions. Following this, the emotion category for each identified emotion span is classified. Finally, the module generates a list of emotion-cause-category triplets, adhering to a specific format to ensure clarity. The workflow also incorporates a formatted triplet output, which facilitates the efficient utilization of data, streamlines further processing, and reduces ambiguity at each stage. This structure ensures that each step in the workflow aligns with the intended task, guiding the model toward accurate and contextually coherent triplet extraction.

% In conclusion, the combination of dialogue-style instructions, rules, and workflows working together effectively empowers large language models to accurately identify emotion and cause spans and categories, thereby successfully fulfilling the defined task objectives.
Consistency between input and output is essential for effective operation. To achieve this, the second dialogue is specifically designed to provide pre-defined, formatted input that adheres to a fixed structure. The model processes this input, and the results are directly generated in the third dialogue as a list of emotion-cause-category triplets. This approach ensures a predictable output that is easy to interpret. In conclusion, the integration of dialogue-style instructions, rules, and workflows ensures that large language models can accurately identify the emotion and cause spans along with their corresponding categories. This combination of strategies enables the model to fulfill the task objectives with precision.

\subsubsection{Low-rank adaptation}
% Direct fine-tuning of large language models, often comprising hundreds of millions to billions of parameters, demands considerable computational resources and time \citep{10.5555/3455716.3455856}, limiting its practicality for many applications. Low-rank adaptation offers a solution by introducing two low-rank matrices, \textit{A} and \textit{B}, and thus adding only a minimal set of trainable parameters to the original model.  This significantly reduces the computational cost, conserves resources, and enhances training efficiency, enabling satisfactory performance even with smaller datasets due to the reduced parameter adjustment overhead.

% To understand LoRA's mechanism, consider the transformation of a standard linear layer's output.  For a linear layer without bias, the output is conventionally:
Optimizing large language models, which typically consist of hundreds of millions to billions of parameters,
presents a significant challenge due to the substantial computational power and time required \citep{10.5555/3455716.3455856}. This complexity makes direct fine-tuning impractical for many applications. To address this issue, low-rank adaptation (LoRA) introduces two compact matrices, denoted as \textit{A} and \textit{B}, which introduce only a limited number of trainable parameters. By leveraging this approach, computational demands are significantly reduced, resource consumption is minimized, and training efficiency is improved. Furthermore, LoRA facilitates effective model adaptation even when working with smaller datasets by limiting the extent of parameter adjustments. To understand the underlying principle of LoRA, it is essential to examine how a standard linear layer processes input. When bias terms are excluded, the output of a typical linear transformation follows the equation:
\begin{align}
&y = Wx
\end{align}
where $W \in \mathbb{R}^{d \times k}$. Here, the input vector $x \in \mathbb{R}^{k}$ represents the latent representation of the input, while $y \in \mathbb{R}^{d}$ represents the output feature vector of the linear layer. $d$ and $k$ correspond to the output and input dimensions of the large language model.
With the integration of LoRA, this formulation is modified to:
\begin{align}
&y = Wx+BAx
\end{align}
% Where $W \in \mathbb{R}^{d \times k}$, $B \in \mathbb{R}^{d \times r}$ and $A \in \mathbb{R}^{r \times k}$. Here, $d$ and $k$ represent the output and input dimensions of the large language model, respectively, and $r$ denotes the rank of the LoRA linear layer ($r \ll d,k$). LoRA fixes the parameters of the original model and only trains the two low-rank matrices $B$ and $A$, thereby achieving efficient fine-tuning.\
In this formulation, $B \in \mathbb{R}^{d \times r}$ and $A \in \mathbb{R}^{r \times k}$. $r$ represents the rank of the LoRA transformation, with the condition ($r \ll d, k$). Rather than updating all parameters of the original model, LoRA retains them in a fixed state and exclusively optimizes the two introduced low-rank matrices, $B$ and $A$. This targeted adjustment significantly enhances fine-tuning efficiency while maintaining computational feasibility.

\section{Experiments}
% In this section, we will elaborate on the experiments specifically designed to evaluate the performance of our proposed model on a range of ECA tasks, encompassing both the span-level emotion-cause-category triplet extraction task and the more fundamental span-level emotion cause analysis (SECA) task. We selected the Chinese emotion cause dataset \citep{GuiWXLZ16}, sourced from Sina City News\footnote{\url{https://news.sina.com.cn/}}. Building upon this, ablation experiments were conducted to delve deeper into the contribution of the instruction design and quantify the impact of key instructional modules.
This section provides a detailed discussion of the experiments designed to assess the effectiveness of the proposed model across various emotion cause analysis tasks. These tasks include both span-level emotion-cause-category triplet extraction and the more fundamental emotion cause span extraction \citep{10.1145/3459637.3482185, li-etal-2021-boundary}. The evaluation was conducted using the Chinese emotion cause dataset \citep{GuiWXLZ16}, which originates from Sina City News\footnote{\url{https://news.sina.com.cn/}}. Furthermore, ablation studies were performed to systematically analyze the influence of the instructional framework and quantify the significance of key instructional components.
To comprehensively evaluate the effectiveness and adaptability of the proposed method across different large language models, two representative LLMs from the current NLP landscape were selected: Baichuan \citep{baichuan2023baichuan2} and GLM \citep{DBLP:conf/acl/DuQLDQY022}. Based on these models, InstruDA\textsubscript{\textit{Baichuan}} and InstruDA\textsubscript{\textit{GLM}} were implemented, representing the application of the proposed approach with Baichuan and GLM, respectively.

\paragraph{\textbf{Implementation details}}
% Our experimental environment consists of PyTorch\footnote{\url{https://pytorch.org/}}, Ubuntu 22.04, and two NVIDIA GeForce RTX 3090 24GB GPUs. Baichuan 2\footnote{\url{https://github.com/baichuan-inc/Baichuan2}} and GLM\footnote{\url{https://open.bigmodel.cn/trialcenter/modeltrial}} are adopted to extract emotion-cause-category triples, while the GLM model generates augmented data for training. The learning rate and the number of training epochs are set to $1 \times 10^{-5}$ and 10, respectively. For evaluation, we implement a ten-fold cross-validation strategy where the dataset is randomly partitioned into ten mutually exclusive subsets. For each validation cycle, nine subsets are used for training while the remaining subset serves as the test partition, with this procedure systematically repeated across all data subdivisions.
The experimental setup is based on PyTorch\footnote{\url{https://pytorch.org/}}, running on Ubuntu 22.04, and utilizes two NVIDIA GeForce RTX 3090 GPUs with 24GB memory. Emotion-cause-category triplet extraction is performed using Baichuan 2\footnote{\url{https://github.com/baichuan-inc/Baichuan2}} and GLM\footnote{\url{https://open.bigmodel.cn/trialcenter/modeltrial}}, with the latter also employed for generating augmented training data. The training process is configured with a learning rate of $1 \times 10^{-5}$ and spans 10 epochs. Model evaluation follows a ten-fold cross-validation approach, where the dataset is randomly divided into ten non-overlapping subsets. In each iteration, nine subsets are used for training while the remaining subset functions as the test set, ensuring that every subset serves as the test partition once during the evaluation process.

\subsection{Comparative methods}
% Given that the SECCE task is still an emerging research area, to address this research gap and establish a performance baseline, we specifically implemented a basic two-step triplet extraction method as our baseline model. To comprehensively evaluate our proposed model, we compared it not only with the aforementioned baseline method but also extensively evaluated it against a range of state-of-the-art information extraction methods.
Considering that the span-level emotion-cause-category triplet extraction task is an emerging area of research, a basic two-step triplet extraction method was developed as a baseline to fill this gap and provide a performance benchmark. To thoroughly assess the effectiveness of our proposed model, comparisons were made not only with this baseline method but also with a variety of state-of-the-art information extraction techniques.
\begin{itemize}
    \item 
    BERT-Classifier: This approach employs a two-step process. Initially, the BERT model is utilized to conduct
sequence labeling for identifying emotion and cause spans within the text. Following this, the labeled emotion
and cause spans are paired together, and a classifier is trained to assign the appropriate emotion category to each
pair.

    \item 
    UIE \citep{lu-etal-2022-unified}: UIE is a unified framework for text-to-structure generation, designed to handle a variety of information extraction tasks.
    
    \item 
    BDTF \citep{zhang-etal-2022-boundary}: BDTF is a table-filling approach that uses a two-dimensional table-based labeling scheme for aspect sentiment triplet extraction. The model is adapted for span-level emotion-cause-category triplet extraction to facilitate comparison with the proposed method.

    \item 
    TF-LaC \citep{YANG2025126062}: TF-LaC is another table-filling method that employs a two-dimensional table-based labeling scheme for emotion-cause-category triplet extraction, leveraging the information from this table representation.
    
\end{itemize}

\subsection{Evaluation metrics}
% We employ span-level precision, recall, and F1-score as evaluation metrics, which are widely used for evaluating span-level emotion cause analysis tasks.
Span-level precision ($P^s$), recall ($R^s$), and F1-score ($F1^s$) are utilized as evaluation metrics, as they are commonly adopted for assessing span-level emotion cause analysis tasks.
\begin{align}
    &P^s = \frac{CT}{PT}\\
    &R^s = \frac{CT}{AT}\\
    &F1^s = \frac{2 \times P^s \times R^s}{P^s + R^s}
\end{align}
% Where $PT$ represents the number of triplets predicted by the model, $CT$ refers to the correct triplets among all predictions, and $AT$ represents the number of annotated triplets in the test set. For span-level evaluation metrics, a triplet is considered correct only if the emotion span, cause span, and emotion category are all predicted correctly. To provide a more comprehensive evaluation of the model, this study further introduces word-level evaluation metrics. These metrics are based on word counts and are less strict compared to span-level metrics. The calculation process is as follows:
where $PT$ denotes the number of triplets predicted by the model, $CT$ represents the correct triplets within those
predictions, and $AT$ indicates the number of annotated triplets in the test set. For span-level metrics, a triplet is deemed correct only if the emotion span, cause span, and emotion category are all accurately predicted. To offer a broader evaluation, this study also incorporates word-level metrics, which are based on word counts and are less stringent than span-level metrics. The calculation process for these word-level metrics is as follows:
\begin{align}
    &P^w = \frac{P^w_{e} + P^w_{c}}{2}\\
    &R^w = \frac{R^w_{e} + R^w_{c}}{2}\\
    &F1^w = \frac{2 \times P^w \times R^w}{P^w + R^w}
\end{align}
where $P^w_{e}$, $R^w_{e}$, $P^w_{c}$ and $R^w_{c}$ are calculated as follows:
\begin{align}
    P^w_{e} = \frac{CE^w}{PE^w},~~~
    R^w_{e} = \frac{CE^w}{AE_w}\\
    P^w_{c} = \frac{CC^w}{PC^w},~~~
    R^w_{c} = \frac{CC^w}{AC^w}
\end{align}
where $CE^w$ and $CC^w$ refer to the number of correctly predicted words within the emotion spans and cause spans of the predicted triplets, respectively. Similarly, $PE^w$ and $PC^w$ represent the total number of words in the predicted emotion and cause spans. On the other hand, $AE^w$ and $AC^w$ indicate the total number of words in the annotated emotion and cause spans in the test set. It is crucial to highlight that for a word in a triplet to be deemed correctly predicted, the corresponding emotion category must also be accurately identified. These word-level evaluation metrics are considered more lenient compared to span-level metrics, as they do not take word order into account and allow for partial matches within spans.

\subsection{Results}
To evaluate the effectiveness of the proposed approach, comprehensive experiments were conducted on two distinct but related tasks: span-level emotion-cause-category triplet extraction and emotion cause span extraction. 

\subsubsection{Results on span-level emotion-cause-category triplet extraction task}

Table~\ref{main results} presents the experimental outcomes, with the highest-performing results emphasized in \textbf{bold}. As shown in
Table~\ref{main results}, the instruction tuning approach, coupled with augmented datasets, significantly outperformed the baseline models. Specifically, the InstruDA\textsubscript{\textit{GLM}} model achieved an 18.86\% improvement in $F1^s$ score over the BERT-Classifier method. When compared to the BDTF method, the InstruDA\textsubscript{\textit{GLM}} model showed a 13.13\% increase in $F1^s$ score. Additionally, in comparison to the UIE method, the InstruDA\textsubscript{\textit{GLM}} model demonstrated a 15.15\% improvement in $F1^s$ score. These results strongly confirm the effectiveness of utilizing instruction tuning and prompt-guided large language models for generating augmented data, thereby boosting model performance. The BERT-Classifier method follows a two-step procedure: first, it identifies all possible emotion and cause spans through sequence labeling, and then it pairs and classifies these pairs in the second step. However, this approach is prone to error propagation, where mistakes made in the first stage are carried over to the second stage, ultimately affecting the accuracy of the extracted triplets. In contrast, the generative method proposed here addresses this issue effectively. Its end-to-end framework minimizes the chances of error accumulation, leading to a notable improvement in the experimental results.
\begin{table}[]
\centering
\setlength{\tabcolsep}{25pt}
\caption{The results on span-level metrics of emotion-cause-category triplet extraction.}
\label{main results}
\begin{tabular}{llll}
\hline
Models  & $P^s$ & $R^s$ & $F1^s$          \\ \hline
BERT-Classifier  & 0.3555     & 0.2963              & 0.3221          \\
UIE    & 0.3594               & 0.3591               & 0.3592         \\
BDTF   & 0.4151               & 0.3493               & 0.3794         \\
TF-LaC & 0.3831               & 0.4546               & 0.4149           \\
\hline
InstruDA\textsubscript{\textit{Baichuan}}   & \textbf{0.5300}              & 0.4875             & 0.5074                \\
% \textbf{ECCT-Bsaichuan} & 0.5148               & 0.4775             & 0.4951           \\
InstruDA\textsubscript{\textit{GLM}} & 0.5268     & \textbf{0.4963}    & \textbf{0.5107}      \\
\hline
\end{tabular}
\end{table}

The InstruDA\textsubscript{\textit{GLM}} model outperformed the two label classification methods, BDTF and TF-LaC. This improved performance can be attributed to instruction tuning's ability to more effectively tap into the vast and varied information embedded in large language models. In contrast, label classification methods are limited by their fixed, predefined label schemas, which restrict their ability to fully harness the rich semantic information within these models. This inherent limitation highlights the advantages of using instruction tuning to guide large language models for triplet generation, resulting in superior performance in text understanding and generation tasks.

A comparative analysis was also conducted between the InstruDA\textsubscript{\textit{GLM}} model and another generative method, UIE. UIE is designed to generate structured extraction language expressions tailored to various information extraction tasks, using specific markers to implement template designs. In contrast, the proposed method utilizes a text-based instruction approach to craft prompt templates. The significant performance boost can be attributed to the richer prompt information provided by text-based instructions, which include task definitions and rule constraints, effectively improving the large language model's understanding of the extraction task.

To further explore the results, a comparative analysis was conducted between the InstruDA\textsubscript{\textit{Baichuan}} and InstruDA\textsubscript{\textit{GLM}} models. As shown in Table~\ref{main results}, both models exhibit notable improvements over the baseline methods, but InstruDA\textsubscript{\textit{GLM}} slightly outperforms InstruDA\textsubscript{\textit{Baichuan}} in the $R^s$ and $F1^s$ metrics. Specifically, InstruDA\textsubscript{\textit{GLM}} shows an approximate 0.88\% higher $R^s$ and about 0.33\% higher $F1^s$ compared to InstruDA\textsubscript{\textit{Baichuan}}. On the other hand, InstruDA\textsubscript{\textit{Baichuan}} exhibits a slightly higher $P^s$, surpassing InstruDA\textsubscript{\textit{GLM}} by around 0.32\%. Despite these small differences, both models perform similarly and significantly outperform other methods. The slight performance advantage of InstruDA\textsubscript{\textit{Baichuan}} could be due to specific architectural features or pre-training data of the Baichuan model, making it marginally more suitable for this task. Nevertheless, the consistent and strong performance of both models emphasizes the effectiveness of the proposed method, regardless of the underlying large language model (Baichuan or GLM) used. In addition to the rigorous span-level evaluation metrics, this research introduces word-level evaluation metrics to offer a more comprehensive evaluation of model performance. As presented in Table~\ref{main results w}, the InstruDA\textsubscript{\textit{Baichuan}} model showed an improvement of 8.49\% in $F1^w$ compared to the two-step BERT-Classifier method. This indicates that the proposed instruction tuning method effectively mitigates error propagation. Moreover, when compared to the UIE model, the InstruDA\textsubscript{\textit{Baichuan}} model achieved a 13.81\% increase in the $F1^w$ score, demonstrating enhanced text comprehension and element identification within triplets. Additionally, increases in the $F1^w$ score of 13.70\% and 7.57\% over the BDTF and TF-LaC models were observed, respectively. These results underscore the generative approach of large language models, which can more flexibly understand and process textual content compared to traditional label classification methods.

\begin{table}[]
\centering
\setlength{\tabcolsep}{25pt}
\caption{The results on word-level metrics of emotion-cause-category triplet extraction.}
\label{main results w}
\begin{tabular}{llll}
\hline
Models  & $P^w$ & $R^w$ & $F1^w$          \\ \hline
BERT-Classifier  & 0.7812     & 0.7541              & 0.7674          \\
UIE    & 0.7088               & 0.7202               & 0.7142         \\
BDTF   & 0.7435               & 0.6871               & 0.7153         \\
TF-LaC & 0.7272               & \textbf{0.8362}               & 0.7766           \\
\hline
InstruDA\textsubscript{\textit{GLM}}     & 0.8284               & 0.7751             & 0.7985          \\
% \textbf{ECCT-Baichuan} & 0.8780               & 0.8014             & 0.8372           \\
InstruDA\textsubscript{\textit{Baichuan}} & \textbf{0.8796}     & 0.8321    & \textbf{0.8523}      \\
\hline
\end{tabular}
\end{table}

Furthermore, a focused comparison between the InstruDA\textsubscript{\textit{Baichuan}} and InstruDA\textsubscript{\textit{GLM}} models reveals a nuanced trend in word-level metrics. Contrary to the previous analysis which suggested a marginal advantage for InstruDA\textsubscript{\textit{Baichuan}}, the Table~\ref{main results w} indicates a more comprehensive performance lead for InstruDA\textsubscript{\textit{Baichuan}}.  Specifically, InstruDA\textsubscript{\textit{Baichuan}} demonstrates superior performance, achieving a higher $P^w$ by approximately 5.12\%, $R^w$ by approximately 5.70\% and $F1^w$ by approximately 5.38\% compared to InstruDA\textsubscript{\textit{GLM}}. This indicates that when considering individual words within the extracted spans, Baichuan is more accurate and comprehensive. On the other hand, the GLM model might have an advantage in covering a broader range of information, enabling it to identify as many potential triplets as possible at the span level, thus achieving a slightly higher performance on span-level metrics.

Overall, these results emphasize that utilizing data augmentation prompts to guide large language models in expanding the training data can substantially improve a model's ability to understand and identify textual information in triplet extraction tasks.

\subsubsection{Results on emotion cause span extraction task}

To further evaluate the performance of our proposed model, experiments were conducted on the emotion cause span extraction task. This task focuses on extracting cause spans from documents that elicit specific emotions. For a comprehensive comparison, the proposed model was benchmarked against several established baseline methods. These include approaches from the work of \cite{li-etal-2021-boundary}, who explored the emotion cause span extraction task as both a sequence labeling and position recognition problem. The sequence labeling methods proposed by \cite{li-etal-2021-boundary}, namely BERT+Softmax, BERT+GRU, and BERT+CRF, utilize BERT to generate contextual embeddings, followed by classifiers such as Softmax, GRU, or CRF to predict labels for each token, identifying the positions of emotion cause spans. In contrast, BERT+Pointer also leverages BERT for feature extraction but employs a pointer network to directly determine the start and end positions of cause spans. Additionally, the proposed model was compared with BGAT, introduced by \cite{10.1145/3459637.3482185}. BGAT is designed to capture the semantic relationship between emotions and text by employing a graph attention network to model the structural information in text, which aims to enhance cause identification accuracy. 
Following the previous emotion cause span extraction works \citep{li-etal-2021-boundary,10.1145/3459637.3482185}, the precision($\hat{P}^s$), recall($\hat{R}^s$), and F1-score ($\hat{F1}^s$) are adopted for evaluation.
% similarly to those used for the span-level emotion-cause-category triplet extraction task, with the primary distinction being that only the extracted cause spans are considered in the evaluation. 

\begin{table}[]
\centering
\setlength{\tabcolsep}{25pt}
\caption{The results on the emotion cause span extraction task.}
\label{SECA_result}
\begin{tabular}{llll}
\hline
Models   & $\hat{P}^s$    & $\hat{R}^s$        & $\hat{F1}^s$              \\ \hline
BERT+Softmax    & 0.4830      & 0.5740      & 0.5250     \\
BERT+GRU & 0.4810          & 0.5670           & 0.5200   \\
BERT+CRF & 0.5640          & 0.5700           & 0.5660   \\
BERT+Pointer & 0.5700         & 0.5260       & 0.5470     \\
BGAT   & 0.7130        & 0.6510        & 0.6790       \\
\hline
InstruDA\textsubscript{\textit{Baichuan}}   & 0.7740     & 0.7360          & 0.7540  \\
InstruDA\textsubscript{\textit{GLM}}    & \textbf{0.8088}           & \textbf{0.7496}          & \textbf{0.7763}       \\
\hline
\end{tabular}
\end{table}

Table~\ref{SECA_result} presents the experimental results of the proposed model and five baseline methods on the emotion cause span extraction task. 
The results demonstrate that the proposed method, incorporating instruction tuning and data augmentation, significantly outperforms the existing approaches.
The InstruDA\textsubscript{\textit{GLM}} model exhibits significant performance improvements over the BGAT model. It achieves enhancements of 9.58\%, 9.86\%, and 9.73\% in $\hat{P}^s$, $\hat{R}^s$, and $\hat{F1}^s$, respectively. These results strongly support the effectiveness of instruction tuning and data augmentation techniques in boosting model performance. Unlike the BGAT model, which utilizes BERT to capture semantic relationships and a graph attention network to model textual structures, the instruction tuning approach in InstruDA\textsubscript{\textit{GLM}} directly directs the model to identify emotion, cause, and category triplets. This method allows the model to better capture and model the complex information present in the text. The observed performance improvement highlights the benefits of using large language models to generate augmented data, which expands the training dataset and leads to further gains in model performance.

A comparative analysis of the performance between the InstruDA\textsubscript{\textit{Baichuan}} and InstruDA\textsubscript{\textit{GLM}} models on the emotion cause span extraction task reveals some subtle, yet significant differences. Moreover, while the performance difference between InstruDA\textsubscript{\textit{GLM}} and InstruDA\textsubscript{\textit{Baichuan}} is relatively small, with InstruDA\textsubscript{\textit{GLM}} showing an approximate 3.48\% higher $\hat{P}^s$, 1.36\% higher $\hat{R}^s$, and 2.23\% higher $\hat{F1}^s$, it consistently demonstrates a marginal advantage. These findings imply that GLM might possess a stronger capacity for modeling span-level information, potentially explaining their improved performance in extracting cause spans. As seen in the span-level emotion-cause-category triplet extraction task, both models exhibit strong and comparable results, significantly outperforming the baseline methods. 
% The slight advantage of InstruDA\textsubscript{\textit{GLM}} on the SECA task suggests potential benefits from the GLM LLM architecture or its specific pre-training for such tasks.

\subsection{Ablation study}

To assess the contribution of key modules within the proposed prompt template, ablation experiments were carried out. In these experiments, the rule module and data augmentation were individually removed in separate trials. The results of these ablation studies are shown in Table~\ref{Ablation_results_Baichuan} and Table~\ref{Ablation_results_GLM}, where ``w/o-Rule-\textit{i}'' refers to the removal of the \textit{i}-th rule and ``w/o-DA'' denotes the exclusion of data augmentation.

\begin{table}[]
\centering
\setlength{\tabcolsep}{25pt}
\caption{The results on the span-level emotion-cause-category triplet extraction task of ablation experiments for InstruDA\textsubscript{\textit{Baichuan}} model.}
\label{Ablation_results_Baichuan}
\begin{tabular}{llll}
\hline
Method   & $P^s$    & $R^s$        & $F1^s$              \\ \hline
InstruDA\textsubscript{\textit{Baichuan}}-w/o-Rule-1    & 0.5220      & 0.4659      & 0.4902     \\
InstruDA\textsubscript{\textit{Baichuan}}-w/o-Rule-2 & 0.5118          & 0.4715           & 0.4904   \\
InstruDA\textsubscript{\textit{Baichuan}}-w/o-Rule-3,4 & 0.5152        & 0.4800           & 0.4966   \\
InstruDA\textsubscript{\textit{Baichuan}}-w/o-Rule-5 & 0.5079         & 0.4560       & 0.4801     \\
InstruDA\textsubscript{\textit{Baichuan}}-w/o-DA  & 0.5148               & 0.4775             & 0.4951           \\
\hline
InstruDA\textsubscript{\textit{Baichuan}}  & \textbf{0.5300}     & \textbf{0.4875}          & \textbf{0.5074}  \\
\hline
\end{tabular}
\end{table}

\begin{table}[]
\centering
\setlength{\tabcolsep}{25pt}
\caption{The results on the span-level emotion-cause-category triplet extraction task of ablation experiments for InstruDA\textsubscript{\textit{GLM}} model.}
\label{Ablation_results_GLM}
\begin{tabular}{llll}
\hline
Method   & $P^s$    & $R^s$        & $F1^s$              \\ \hline
InstruDA\textsubscript{\textit{GLM}}-w/o-Rule-1    & 0.5219      & 0.4854      & 0.4996     \\
InstruDA\textsubscript{\textit{GLM}}-w/o-Rule-2 & 0.5280          & 0.4922           & 0.5093   \\
InstruDA\textsubscript{\textit{GLM}}-w/o-Rule-3,4 & 0.5268        & 0.4935           & 0.5094   \\
InstruDA\textsubscript{\textit{GLM}}-w/o-Rule-5 & 0.5153         & 0.4843            & 0.4991     \\
InstruDA\textsubscript{\textit{GLM}}-w/o-DA     & 0.5053               & 0.4864             & 0.4955           \\
\hline
InstruDA\textsubscript{\textit{GLM}}  & \textbf{0.5268}     & \textbf{0.4963}          & \textbf{0.5107}  \\
\hline
\end{tabular}
\end{table}
\paragraph{Impact of rules.} To gain deeper insights into the influence of the rule module, individual ablation experiments were conducted for each rule. Specifically, \textit{Rule 3} and \textit{Rule 4}, which both help the model comprehend the inherent relationships between emotions, causes, and categories, were removed together in a single experiment. 
% The performance variations from these ablation experiments are presented in Table~\ref{Ablation_results_Baichuan} and Table~\ref{Ablation_results_GLM}, highlighting the impact of excluding these rules on the overall results. 
From Table \ref{Ablation_results_Baichuan} and Table \ref{Ablation_results_GLM}, it can be observed that the removal of different rules leads to distinct performance variations, reflecting their unique roles:
(1) The exclusion of \textit{Rule 1}: As shown in Table~\ref{Ablation_results_Baichuan}, the exclusion of \textit{Rule 1} (denoted as ``w/o-Rule-1'') results in a noticeable decline in performance. Specifically, for $P^s$ decreases by 0.80\%, $R^s$ drops by 2.16\%, and $F1^s$ declines by 1.72\% when compared to the full InstruDA\textsubscript{\textit{Baichuan}} model. Similarly, for the InstruDA\textsubscript{\textit{GLM}} model (Table~\ref{Ablation_results_GLM}), removing \textit{Rule 1} also leads to a performance decrease. This reduction in performance underscores the important role that \textit{Rule 1} plays in sustaining model efficacy. \textit{Rule 1} is intended to ensure the model strictly follows user instructions and stays within the defined workflow. Its removal diminishes the model’s compliance with these instructions, which can lead to inconsistencies or deviations in task execution.
(2) The removal of \textit{Rule 2}: The data in Tables~\ref{Ablation_results_Baichuan} and~\ref{Ablation_results_GLM} indicate that the removal of \textit{Rule 2} (denoted as ``w/o-Rule-2'') results in a slight decline in model performance, with $P^s$ dropping by 1.82\%, $R^s$ by 1.60\%, and $F1^s$ by 1.70\%. Likewise, for InstruDA\textsubscript{\textit{GLM}}, a slight decrease in performance is observed, with $F1^s$ reducing by 0.14\%, suggesting a similar role for \textit{Rule 2} in maintaining the integrity of extracted spans for the GLM model. Although the performance decline is smaller than that observed with the removal of \textit{Rule 5}, it still highlights the positive impact of \textit{Rule 2} on model effectiveness. \textit{Rule 2} serves to prevent the model from modifying the original text during the extraction process, thus avoiding non-compliant spans, such as cause spans extending beyond clause boundaries. When \textit{Rule 2} is excluded, the model faces fewer restrictions on text manipulation, potentially leading to unintended alterations during span extraction. Although these changes may not always be semantically erroneous, they can introduce textual inconsistencies, negatively affecting evaluation metrics, particularly those relying on precise span-level matching.
(3) The removal of \textit{Rule 3} and \textit{Rule 4}: The results demonstrate that the removal of \textit{Rule 3} and \textit{Rule 4} (denoted as ``w/o-Rule-3,4'') results in a noticeable decline in model performance. Specifically, $P^s$ decreases by 1.48\%, $R^s$ by 0.75\%, and $F1^s$ by 1.08\% compared to the full InstruDA\textsubscript{\textit{Baichuan}} model. For InstruDA\textsubscript{\textit{GLM}}, a comparable performance drop is seen, with $F1^s$ decreasing by 0.13\%. This reduction underscores the pivotal role of \textit{Rule 3} and \textit{Rule 4} in assisting the model to effectively leverage the inherent relationships between emotion spans, cause spans, and emotion categories. The performance drop suggests that these rules are crucial for enabling the model to accurately combine potential spans into coherent and valid emotion-cause-category triplets.
(4) The removal of \textit{Rule 5}: As verified in both ablation tables, the removal of \textit{Rule 5} (denoted as "w/o-Rule-5") results in a more significant performance decline compared to the removal of \textit{Rule 3} and \textit{Rule 4}. Specifically, $P^s$ decreases by 2.21\%, $R^s$ by 3.15\%, and $F1^s$ by 2.73\%. Similarly, InstruDA\textsubscript{\textit{GLM}} also exhibits a performance reduction, with $F1^s$ dropping by 1.52\%. This notable reduction in performance highlights the essential role of \textit{Rule 5} in ensuring the generation of contextually coherent and textually grounded triplets.

\paragraph{Impact of data augmentation.} 
A direct comparison between ``InstruDA\textsubscript{\textit{Baichuan}}-w/o-DA'' and InstruDA\textsubscript{\textit{Baichuan}} highlights the advantages of incorporating data augmentation. InstruDA\textsubscript{\textit{Baichuan}}, which includes data augmentation, consistently surpasses ``InstruDA\textsubscript{\textit{Baichuan}}-w/o-DA'' in all evaluation metrics. Specifically, InstruDA\textsubscript{\textit{Baichuan}} shows improvements of 1.52\% in $P^s$, 2.00\% in $R^s$, and 1.23\% in $F1^s$ over ``InstruDA\textsubscript{\textit{Baichuan}}-w/o-DA''. These improvements may be attributed to the expanded diversity and volume of training data resulting from data augmentation. By exposing the model to a broader range of textual variations and emotional expressions, data augmentation enhances InstruDA\textsubscript{\textit{Baichuan}}’s ability to generalize, improving its robustness and accuracy in triplet extraction tasks. Likewise, for the InstruDA\textsubscript{\textit{GLM}} model, comparing ``InstruDA\textsubscript{\textit{GLM}}-w/o-DA'' with InstruDA\textsubscript{\textit{GLM}} shows that data augmentation is also beneficial. 

\section{Case study}
The case studies are based on the InstruDA\textsubscript{\textit{Baichuan}} model to illustrate the practical implications of rule and workflow modules. Both the rule module and the workflow module are critical to the model’s effective operation. In the absence of either module, the model fails to generate the predefined structured output, and more importantly, the semantic coherence and overall quality of the emotion-cause-category triplets degrade considerably. To explore the reasons behind this performance drop, a thorough analysis of prediction patterns is conducted, supported by illustrative case studies.

\subsection{Impact of rule module}

As previously discussed, the rule module is designed to ensure strict adherence to the task’s requirements and steer the model toward accurate triplet extraction. Comprising five specific rules, this module addresses various facets of controlled, contextually relevant output. To directly evaluate the impact of the rule module on overall performance, Table \ref{cases} presents two experimental scenarios illustrating the model’s predictions when the rule module is omitted. In Case 1, although the model generates text in the correct triplet format without the rule module, the content of the triplet is incorrectly predicted. In Case 2, the model fails to produce text that aligns with the desired triplet format altogether. In Case 1, the identified emotion span is ``happy'' with the corresponding cause span being ``this child has become a part of our family'' categorized under the emotion type ``Happiness''. As illustrated in Table \ref{cases}, the complete model successfully predicts the correct triplet. However, when the rule module is removed, the model erroneously selects ``seeing the child’s face turning purple from the cold'' as the cause span. This misclassification likely arises due to the absence of \textit{Rule 3} and \textit{Rule 5}, which play a crucial role in ensuring precise alignment between emotion spans and their associated cause spans. In particular, \textit{Rule 3} mandates that the model correctly recognize both spans and establish their proper correspondence. Without this guiding constraint, the model incorrectly links ``happy'' to ``seeing the child’s face turning purple from the cold'' instead of ``this child has become a part of our family'', thereby failing to capture the true semantic relationship between the emotion and its cause.
% As detailed previously, a rule module is incorporated to enforce strict adherence to the task definition and guide the model toward accurate triplet extraction. This module is composed of five rules aimed at various aspects of the controlled and contextually relevant output. To directly assess the contribution of this rule module to overall performance, Table~\ref{cases} presents two experimental cases that reflect the model's predictions after the rule module is removed. In Case 1, although the model without the rule module generates text that follows the pre-defined triplet format, the content of the triplet is incorrectly predicted. In Case 2, the model fails to generate text that conforms to the triplet format.

\begin{table}[!ht]\small
 \caption{Two cases selected from the dataset, where ``w/o-Rules'' indicates predictions from the model with the rule module removed, and ``InstruDA\textsubscript{\textit{Baichuan}}'' indicates predictions from the full model.}
\label{cases} 
\begin{center}
\begin{tabular}{p{14.0cm}}
\hline
\textbf{Case 1}: This year, 43-year-old Ani Kurban, a forest ranger, grew up in poverty. His father passed away from illness when he was six months old, and his mother raised him and his four brothers alone. He recalled: ``On the morning of March 5, 2003, around 6 a.m., my wife went out to sweep the streets and found a small basket on the roadside, wrapped in a floral cloth with a child inside and a note beside it. Seeing this poor little life, my wife carefully brought the child home and discussed it with me. Seeing the child's face turning purple from the cold, I quickly wrapped the child in a blanket to keep them warm. Since then, [\textit{Cause} 1]\textless this child has become a part of our family\textgreater, and we are [\textit{Emotion} 1]\{happy\}.'' \\
\textbf{[Gold triplets] - (happy, this child has become a part of our family, Happiness)} \\
\textbf{[w/o-Rules] - (happy, seeing the child's face turning purple from the cold, Happiness)}   \\
\textbf{[InstruDA\textsubscript{\textit{Baichuan}}] - (happy, this child has become a part of our family, Happiness)}  \\
\hline
\textbf{Case 2}: Grandma Ling, 86 years old, can cook and do laundry by herself, but she is [\textit{Emotion} 1]\{afraid\} [\textit{Cause} 1]\textless of spending the night alone\textgreater. ``As soon as it gets dark, I look forward to Xiao Yun coming over''. The old lady's home is a small two-bedroom apartment, but it looks very clean. ``This is all thanks to Xiao Yun. She always helps clean up when she comes over at night, boiling water and tidying up, making the house very neat''. \\
\textbf{[Gold triplets] - (afraid, spending the night alone, Fear)} \\
\textbf{[w/o-Rules] - (0. afraid of spending the night alone, 'Fear', 'As soon as it gets dark' 1. very happy, 'Happiness', 'This is all thanks to Xiao Yun')}   \\
\textbf{[InstruDA\textsubscript{\textit{Baichuan}}] - (afraid, spending the night alone, Fear)}  \\
\hline
\end{tabular}
\end{center}
\vspace{-0.5em}
\end{table}

\textit{Rule 5} ensures that the model incorporates contextual understanding when extracting triplets, thereby maintaining coherence and logical consistency within the text. Without this rule, the model struggles to effectively utilize surrounding information to validate the plausibility of its predictions, resulting in deviations from accurate triplet extraction. Essentially, \textit{Rule 5} plays a critical role in reinforcing contextual grounding, allowing the model to verify the legitimacy of the extracted triplets. Its absence weakens the model’s ability to reason contextually, leading to inconsistencies in predictions. Furthermore, the lack of \textit{Rule 1}, which enforces strict adherence to predefined extraction guidelines, exacerbates this issue by further impairing the model’s reasoning framework. This combined deficiency ultimately contributes to the erroneous triplet prediction observed in Case 1.

In Case 2, the identified emotion span is ``afraid'', with its corresponding cause span being ``spending the night alone'' categorized under the ``Fear'' emotion type. As illustrated in Table \ref{cases}, the complete model accurately predicts the triplet. However, once the rule module is removed, the model exhibits a substantial decline in performance, failing not only to maintain the prescribed triplet format but also to generate semantically meaningful triplets. This deterioration is directly linked to the absence of \textit{Rule 1}, which enforces strict compliance with structured extraction guidelines. Without this rule, the model is unable to produce outputs that conform to the predefined format. A closer examination reveals that the model without the rule module extracts three spans: ``afraid of spending the night alone'', ``As soon as it gets dark'', and ``This is all thanks to Xiao Yun''. The key issue here is that it incorrectly merges the emotion and cause spans, treating ``afraid of spending the night alone'' as a single entity instead of distinguishing ``afraid'' as the emotion and ``spending the night alone'' as the cause. This failure highlights a significant shortcoming in the model’s ability to decompose text into distinct emotion and cause spans, a process governed by \textit{Rule 3}. Additionally, the erroneous extraction of unrelated spans, such as ``As soon as it gets dark'' and ``This is all thanks to Xiao Yun'' further underscores the model’s weakened ability to accurately identify relevant spans once \textit{Rule 3} is no longer guiding its predictions.

Additionally, the model erroneously identifies ``very happy'' as the emotion span for the event text ``This is all thanks to Xiao Yun'', despite no such emotion being present in the original text. This fabricated output underscores the importance of \textit{Rule 2}, which is specifically designed to prevent the generation of hallucinated content. In the absence of \textit{Rule 2}, the model relies on its own potentially flawed inferencing, leading to the inclusion of unsubstantiated information. This not only compromises the accuracy of the extracted triplets but also results in a deviation from the factual basis of the text, thereby undermining the reliability of the model’s predictions.

In summary, the removal of the rule module eliminates crucial constraints, leading to persistent failures in the model’s ability to generate accurate and properly formatted triplets. These inaccuracies manifest as violations of the standard triplet structure or the introduction of content that lacks textual support. This highlights the essential role of the rule module in maintaining both the structural coherence and factual accuracy of the model’s predictions.

\subsection{Impact of workflow module}

The workflow module operationalizes the methodological framework by systematically guiding text processing and structuring triplet extraction. It follows a structured sequence: identifying emotion spans, extracting corresponding cause spans, classifying emotion categories, and formatting the final triplet output. To assess the significance of this structured process and examine the impact of its removal on model performance, Table \ref{cases2} presents two experimental cases illustrating the model’s predictions without the workflow module. In Case 3, the emotion span is identified as ``moved'' with ``Jiang Xingbao’s act of bravery'' as the corresponding cause span and the emotion category classified as ``Happiness''. As shown in Table \ref{cases2}, the full model correctly predicts the triplet. However, when the workflow module is removed, the model not only fails to generate output in the prescribed triplet format but also misclassifies the emotion category, incorrectly associating ``moved'' with ``Sadness''. This highlights the essential role of the workflow module in providing structured, step-by-step guidance, ensuring both accurate emotion classification and the proper formatting of extracted triplets.

\begin{table}[!ht]\small
 \caption{Two cases selected from the dataset, where ``w/o-Workflow'' indicates predictions from the model with the workflow module removed, and ``InstruDA\textsubscript{\textit{Baichuan}}'' indicates predictions from the full model.}
\label{cases2} 
\begin{center}
\begin{tabular}{p{14.0cm}}
\hline
\textbf{Case 3}: [\textit{Cause} 1]\textless Jiang Xingbao's act of bravery\textgreater, the neighbors around him saw it and heard about it, and everyone felt [\textit{Emotion 1}]\{moved\}. Jiang Xingbao's neighbor said: ``Jumping from such a high building, if he hadn't caught him, this person would have... Even if someone fell from the second floor, we wouldn't dare to catch them, let alone from the fourth floor. He was really brave.'' \\
\textbf{[Gold triplets] - (moved, Jiang Xingbao's act of bravery, Happiness)} \\
\textbf{[w/o-Workflow] - (1. moved, Jiang Xingbao's act of bravery, Sadness)}   \\
\textbf{[InstruDA\textsubscript{\textit{Baichuan}}] - (moved, Jiang Xingbao's act of bravery, Happiness)}  \\
\hline
\textbf{Case 4}: In June of this year, after Mr. Su's elderly father passed away due to illness, his aunt revealed the truth, saying that his mother was living in Cangzhou, but they had not been in contact for over 30 years, and her current situation was unknown. [\textit{Cause} 1]\textless Learning that his biological mother might still be alive\textgreater, Mr. Su could not hide his [\textit{Emotion 1}]\{excitement\}. Following the clues provided by his aunt, he went to Cangzhou alone to search for his mother. His aunt said that his mother was from Nanpi County, so Mr. Su first went to the Nanpi County Civil Affairs Bureau to inquire. The staff told him that his mother had previously worked at the Cangzhou First Vocational School and might be living in Cangzhou. \\
\textbf{[Gold triplets] - (excitement, learning that his biological mother might still be alive, Happiness)} \\
\textbf{[w/o-Workflow] - (1. excitement, learning that his biological mother might still be alive, Mr. Su could not hide his excitement, excitement, Sadness)}   \\
\textbf{[InstruDA\textsubscript{\textit{Baichuan}}] - (excitement, learning that his biological mother might still be alive, Happiness)}  \\
\hline
\end{tabular}
\end{center}
\vspace{-0.5em}
\end{table}

% In summary, removing the workflow module, designed to methodically guide triplet generation, demonstrably impairs performance. The resulting outputs frequently fail to adhere to the expected triplet format and misclassify emotion categories. This clearly illustrates that a well-defined workflow is not merely beneficial, but a critical prerequisite for ensuring the integrity of our experimental evaluations. Consequently, the workflow module is vital for generating reliable and rigorously validated findings.
In Case 4, the identified emotion span is ``excitement'', with the corresponding cause span being ``learning that his
biological mother might still be alive'' and the emotion category classified as ``Happiness''. As indicated in Table \ref{cases2}, the full model accurately predicts the triplet. However, in the absence of the workflow module, the model once again fails to adhere to the predefined format and repeats the misclassification error, incorrectly categorizing ``excitement'' under ``Sadness''. An analysis of the output suggests that even without the workflow module, the model retains a certain level of semantic processing, allowing it to recognize sentences containing emotional expressions and extract the emotion span ``excitement''. However, its ability to precisely identify the corresponding cause span and organize the extracted elements into properly formatted triplets is significantly impaired. This pattern indicates that while some degree of semantic comprehension persists, the workflow module plays a crucial role in providing the structured algorithmic framework necessary for transforming raw text and partial semantic understanding into consistently formatted and accurate emotion-cause-category triplets. In conclusion, the removal of the workflow module, which is designed to systematically guide the triplet generation process, significantly hampers the model’s performance. The outputs often fail to conform to the required triplet format and misclassify emotion categories. This highlights the necessity of a well-structured workflow, not just as a useful tool but as an essential component for maintaining the integrity of our experimental results. Therefore, the workflow module plays a crucial role in ensuring the reliability and rigor of the findings.

\section{Conclusion and future work}
% In conclusion, we have presented a novel instruction tuning approach, coupled with large language model-driven data augmentation, to effectively address the complex span-level emotion-cause-category triplet extraction task. By crafting targeted triplet extraction instructions, we successfully harnessed the potent language understanding and reasoning abilities of large language models to generate pre-defined formatted triplets. Furthermore, the integration of low-rank adaptation efficiently reduces the computational burden of fine-tuning. Data augmentation prompts, designed to guide large language models in producing supplementary data, effectively expand the training data and enhance model robustness. The experimental findings definitively demonstrate the significant advantage of our proposed method.
In conclusion, this study has presented a novel instruction tuning approach, integrated with large language model-driven data augmentation, to effectively address the intricate task of span-level emotion-cause-category triplet extraction. By crafting targeted triplet extraction instructions, the inherent language understanding and reasoning abilities of large language models were successfully harnessed to generate pre-defined formatted triplets. The incorporation of low-rank adaptation further mitigated the computational burden of fine-tuning. Moreover, data augmentation prompts, designed to guide large language models in the production of supplementary data, effectively expanded training resources and enhanced model robustness. The experimental findings definitively demonstrate the significant advantages offered by the proposed methodology.

Future research directions should focus on optimizing data generation strategies for complex structured prediction tasks. While this work employed context replacement for data augmentation, more sophisticated generation mechanisms could further enhance triplet quality and diversity. The development of systematic frameworks for automated large-scale triplet annotation using advanced text generation capabilities remains crucial for advancing emotion analysis and related fields. Additionally, exploring dynamic instruction adaptation mechanisms and cross-task transfer learning could strengthen model generalization across different structured prediction scenarios. These advancements would substantially reduce manual annotation efforts while improving the precision and applicability of emotion cause analysis systems.
\bibliographystyle{cas-model2-names}
\bibliography{ref}
% \bibliography{ref}
\end{document}